\pdfoutput=1

\documentclass[11pt]{article}

\usepackage[final]{acl}
\usepackage[most]{tcolorbox}

\newtcolorbox{examplebox}[1]{%
  enhanced,
  colback=gray!8,          
  colframe=gray!55,        
  coltitle=white,          
  colbacktitle=gray!90,    
  fonttitle=\bfseries,
  title=#1,
  boxrule=0.7pt,
  arc=3pt,
  left=8pt,right=8pt,top=6pt,bottom=6pt,
  boxed title style={%
    boxrule=0pt,
    sharp corners,
    interior style={fill=gray!90},
    size=fbox,             
    left=8pt,right=8pt,top=3pt,bottom=3pt
  }
}

\usepackage{times}
\usepackage{latexsym}
\usepackage[table]{xcolor}
\definecolor{lightwine}{RGB}{255, 235, 238} 
\definecolor{lightlavender}{RGB}{237, 237, 255} 
\usepackage{afterpage}
\usepackage{enumitem}

\usepackage[T1]{fontenc}

\usepackage[utf8]{inputenc}

\usepackage{microtype}

\usepackage{inconsolata}

\usepackage{graphicx}
\usepackage{booktabs} 
\usepackage{amsmath}
\usepackage{algorithm}
\usepackage{algpseudocode}
\usepackage{placeins}

\newcounter{example}
\renewcommand{\theexample}{\arabic{example}}

\usepackage{dblfloatfix}

\title{
AdaFuse: Adaptive Ensemble Decoding with Test-Time Scaling for LLMs}

\author{
\textbf{Chengming Cui}\textsuperscript{1†}\quad
\textbf{Tianxin Wei}\textsuperscript{1†}\quad
\textbf{Ziyi Chen}\textsuperscript{1}\quad
\textbf{Ruizhong Qiu}\textsuperscript{1}\quad
\textbf{Zhichen Zeng}\textsuperscript{1}
\\
\textbf{Zhining Liu}\textsuperscript{1}\quad
\textbf{Xuying Ning}\textsuperscript{1}\quad
\textbf{Duo Zhou}\textsuperscript{1}\quad
\textbf{Jingrui He}\textsuperscript{1}
\\
\textsuperscript{†}Equal contribution \\
\textsuperscript{1}University of Illinois Urbana-Champaign
\\
\texttt{\{ccui12, twei10, jingrui\}@illinois.edu}
}

\begin{document}
\maketitle
\begin{abstract}

Large language models (LLMs) exhibit complementary strengths arising from differences in pretraining data, model architectures, and decoding behaviors. Inference-time ensembling provides a practical way to combine these capabilities without retraining. However, existing ensemble approaches suffer from fundamental limitations. Most rely on fixed fusion granularity, which lacks the flexibility required for mid-generation adaptation and fails to adapt to different generation characteristics across tasks. To address these challenges, we propose \textsc{AdaFuse}, an adaptive ensemble decoding framework that dynamically selects semantically appropriate fusion units during generation. Rather than committing to a fixed granularity, \textsc{AdaFuse} adjusts fusion behavior on the fly based on the decoding context, with words serving as basic building blocks for alignment. To be specific, we introduce an uncertainty-based criterion to decide whether to apply ensembling at each decoding step. Under confident decoding states, the model continues generation directly. In less certain states, \textsc{AdaFuse} invokes a diversity-aware scaling strategy to explore alternative candidate continuations and inform ensemble decisions. This design establishes a synergistic interaction between adaptive ensembling and test-time scaling, where ensemble decisions guide targeted exploration, and the resulting diversity in turn strengthens ensemble quality. Experiments on open-domain QA, arithmetic reasoning, and machine translation demonstrate that \textsc{AdaFuse} consistently outperforms strong ensemble baselines, achieving an average relative improvement of 6.88\%. The code is available at \url{https://github.com/CCM0111/AdaFuse}.

\end{abstract}

\section{Introduction}

Large language models (LLMs) have demonstrated strong performance across a wide range of natural language processing tasks \cite{hendrycks2020measuring,cobbe2021training,ning2025mc,wei2025evo,ai2025resmoe,zhang2025improving,chen2024wapiti} and are increasingly deployed in real-world applications \cite{grattafiori2024llama,openai2024gpt4}.
However, model performance is not uniform across tasks: differences in pretraining data, model architectures, and decoding strategies lead individual models to exhibit heterogeneous strengths\cite{chu2025think}.
Models that excel at structured or multi-step reasoning may perform less reliably on open-domain or commonsense question answering, while models optimized for such queries may struggle with complex reasoning
\cite{joshi2017triviaqa,kwiatkowski2019natural,huang-chang-2023-towards,guo2025deepseek}.
As a result, inference-time ensembling has emerged as a practical approach for improving robustness by combining complementary model strengths without retraining \cite{jiang2023llm,liu2024cool,yao2024determine,wang2023selfconsistencyimproveschainthought}.

Existing inference-time ensemble methods differ primarily in the granularity at which model outputs are fused \cite{chen2023frugalgptuselargelanguage}.
Sample-level approaches combine complete model responses via reranking or response fusion \cite{jiang2023llm,shnitzer2023large,lu2023routing,wang2023fusing}, but perform fusion only after generation has finished, preventing any mid-generation correction.
Span-level methods assemble multi-token segments produced by different models \cite{liu2024cool,xu2024hit,lv2024specfuse,fu2025rlae}, yet rely on predefined or fixed span boundaries that limit flexibility across tasks and reasoning depths.
Token-level approaches aggregate next-token distributions at each decoding step \cite{yao2024determine,huang2024ensemble,yu2024breaking,zeng2025harnessing}, but typically require token alignment across models, restricting applicability under heterogeneous tokenizers; related byte-level alternatives alleviate this mismatch at the cost of increased computation and weakened semantic structure \cite{phan2024exact,sathe2024improving}.
Despite operating at different granularities, most existing methods rely on a \emph{fixed} fusion resolution, which restricts their ability to adapt the fusion scope to evolving semantic context, task demands, and reasoning uncertainty during generation \cite{yao2023tree}.

These trade-offs motivate the need for an adaptive inference-time ensemble framework that overcomes the limitations imposed by fixed-granularity generation.
We adopt word-level units as the basic building blocks, as they provide a natural abstraction that supports step-wise integration and semantic coherence, while avoiding explicit token alignment across heterogeneous tokenizers.
Based on this design choice, we propose \textsc{AdaFuse}, an adaptive word-level ensembling framework that enables flexible inference-time fusion and mid-generation correction at natural word boundaries, which preserves semantic integrity across models.

Concretely, \textsc{AdaFuse} adopts an adaptive decoding strategy that balances effectiveness and efficiency during generation. We introduce an uncertainty-based criterion to decide whether to apply ensembling at each decoding step. Under confident decoding states, the model proceeds with direct generation. In less certain states, \textsc{AdaFuse} invokes a diversity-aware scaling strategy to explore alternative candidate continuations and support more effective ensemble decisions. This design is guided by the evolving decoding context rather than predefined rules, and establishes a synergistic interaction between adaptive ensembling and test-time scaling. In particular, selective exploration under uncertainty improves test-time scaling effectiveness, while the resulting diversity further strengthens ensemble quality.

In summary, our paper contributes the following:

\begin{itemize}[itemsep=-0.2em, topsep=-0.3em, leftmargin=*]

    \item \textbf{Confidence-Guided Adaptive Decoding}:
    \textsc{AdaFuse} uses model confidence to decide when to commit longer word spans
    and when to consider more candidate continuations, enabling adaptive mid-generation
    correction without relying on fixed-length spans.

    \item \textbf{Diversity-Aware Ensemble Scaling}:
    Building on confidence-guided decoding, \textsc{AdaFuse} employs an exploration strategy to generate diverse candidate continuations for ensemble scoring only when uncertainty arises, while avoiding unnecessary computation in confident states.
    

    \item \textbf{Strong Empirical Performance Across Diverse Tasks}:
    \textsc{AdaFuse} achieves consistent gains over the strongest ensemble baselines, with an average relative improvement of 6.88\% across open-domain question answering, arithmetic reasoning, and machine translation benchmarks.

\end{itemize}

\section{Related Works}
Ensemble methods in LLMs have evolved significantly, with prominent approaches categorized into three main levels: \textit{Sample-level, Span-level, Token-level ensembling}, each designed to harness complementary model strengths yet plagued by its own drawbacks.

\paragraph{Sample‑Level Ensembling} aggregates entire generated responses from multiple LLMs. LLM‑Blender \cite{jiang2023llm} trains a \emph{pairwise ranker} to compare and select the strongest candidate, then applies a \emph{generative fusion} module to produce a final answer. Routing methods \cite{shnitzer2023large,lu2023routing} assign each query to the most suitable expert model, but their effectiveness is capped when every candidate output contains errors \cite{hu2024routerbench}. Fusion networks \cite{wang2023fusing} go one step further by learning to merge several complete outputs into a single consolidated response, boosting overall quality but still facing challenges in generalizing across diverse tasks and overlooking valuable token‑level probability cues generated during decoding \cite{lin2025life,li2025smoa}.
\paragraph{Span-Level Ensembling} combines output segments. Cool-Fusion~\cite{liu2024cool} merges model outputs once they reach common word boundaries, while SweetSpan~\cite{xu2024hit} generates and merges spans based on perplexity scores. SpecFuse~\cite{lv2024specfuse} aggregates multiple LLM outputs by predicting the next segment and combining them. RLAE \cite{fu2025rlae} uses reinforcement learning to adjust weights dynamically during the merging process. While span-level methods improve coherence and fluency, they rely on the quality of the generated segments, making them susceptible to errors in weak segments. Furthermore, span-level methods offer limited flexibility due to their fixed fusion granularity and lack of adaptive control, making them less effective on novel or diverse inputs.

\begin{figure*}[t]
    \centering
    \includegraphics[width=1\textwidth]{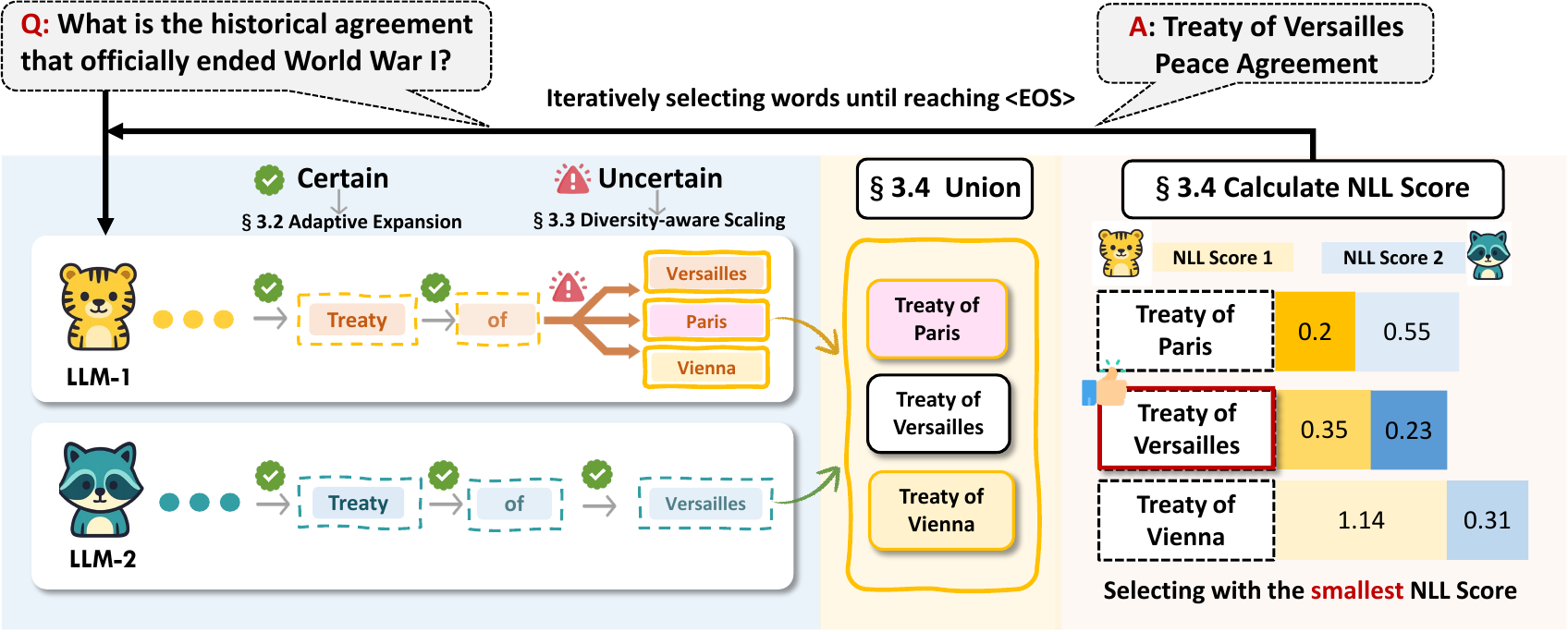}
    \caption{
    Illustration of the \textsc{AdaFuse} framework. AdaFuse ensembles multiple LLMs during decoding by dynamically adjusting fusion behavior. It expands in confident decoding states and applies diversity-aware scaling under uncertainty. Candidate word sequences are unified and selected using NLL scores, enabling adaptive and stable generation.
    }
    \label{fig:overview}
\end{figure*}

\paragraph{Token‑Level Ensembling} merges each model's next‑token distributions at every decoding step. UniTE \cite{yao2024determine} forms the union of each model's top‑k token candidates at each step, avoiding full‑vocabulary alignment and cutting inference cost. DeePEn \cite{huang2024ensemble} projects each model's logits into a universal relative space via anchor‑token similarity, aggregates them, and applies a search‑based inverse transformation to recover logits in a main model's vocabulary, enabling training‑free fusion of heterogeneous LLMs. GAC \cite{yu2024breaking} builds dense mapping matrices over the union vocabulary, projects each model's probability vector into this joint space, and averages them to select the next token, optionally cascading to cheaper ensembling when confidence is high. Despite their effectiveness, token-level approaches operate over large vocabularies at every step, leading to substantial computational overhead. They also struggle with heterogeneous tokenizations and may overlook low-probability yet semantically important tokens. Byte-level ensembling addresses vocabulary mismatch by merging distributions at the byte granularity \cite{phan2024exact}, but introduces constant per-step conversion overhead and obscures word-level semantic structure. These limitations motivate our adaptive ensembling framework, which performs confidence-guided fusion at natural word boundaries, enabling precise step-wise integration with reduced overhead and improved semantic alignment.




\section{Methodology}

We present \textsc{AdaFuse}, an adaptive ensemble decoding framework that addresses a core limitation of existing inference-time ensembling methods: their reliance on fixed fusion granularity and rigid decoding control, which prevents fusion behavior from adapting to evolving uncertainty.

At each decoding step, \textsc{AdaFuse} adaptively generates word-level candidate segments from each model, regulates the breadth of candidate consideration based on model confidence, and applies joint scoring across models to select the most appropriate continuation. 
This design enables real-time integration of ensemble preferences at a linguistically meaningful granularity, while flexibly balancing robustness and efficiency during decoding. 
An overview is shown in \autoref{fig:overview}.

\subsection{Candidate Word Proposal}
\label{sec:candidate_word_proposal}

\textsc{AdaFuse} generates candidate word completions from each model by extending the current decoding prefix until a word boundary is reached. Starting from the current prefix $\mathbf{y}_{1:t}$, each model $m_k$ generates tokens sequentially according to its next-token distribution until a complete word is formed. The resulting token sequence is decoded into a word segment, denoted as $\mathbf{w}$. This design avoids mid-word truncation and provides a consistent unit for alignment and scoring across models during ensemble decoding.






\subsection{Adaptive Word Commitment}
\label{subsec:adaptive-word-commitment}

In word-level ensemble decoding, we need an effective rule to decide whether the model is sufficiently confident to continue generating a longer span or to stop and re-score candidate segments. Because a full word or phrase typically spans multiple subword tokens, its confidence cannot be reliably assessed before generation completes. We therefore use the first-token signal as a practical proxy for how ready the model is to commit to extending the current word.

\paragraph{Start-of-Word Confidence.}
We assess confidence using the margin between the top-ranked candidates in the first-token distribution at the start of a word. 
Specifically, let $p_{(1)}$ and $p_{(2)}$ be the top-1 and top-2 probabilities of the \emph{first token} at the start of a word. We define the confidence margin as $\Delta_{m_k}(\mathbf{y}) = p_{(1)} - p_{(2)}$.

A larger margin corresponds to higher confidence, allowing the decoder to continue generating a longer word span, whereas a smaller margin indicates uncertainty and triggers early re-scoring.

\paragraph{Adaptive Expansion Strategy.}
Based on the confidence signal defined above, \textsc{AdaFuse} adaptively decides
whether to extend generation beyond the current word or to stop and perform
ensemble re-scoring.
For model $m_k$ at prefix $\mathbf{y}$, we allow continued generation only when the
margin criterion exceeds a predefined threshold $\tau_\Delta$:
\begin{equation}
\label{eq:confidence-criterion}
\Delta_{m_k}(\mathbf{y}) \ge \tau_\Delta.
\end{equation}
When this condition is satisfied, the decoder proceeds to continue generating the current word
using greedy decoding, treating the current continuation as a confident span
extension.
Otherwise, generation is halted after the current word and candidate segments
are returned for cross-model scoring.

To avoid over-commitment while preserving step-wise correction, we limit each decoding round to at most $M$ generated words.
In practice, we set $M=3$ in all experiments, which is sufficient to capture most compact semantic units (e.g., named entities, short phrases, and collocations), while preventing error accumulation and delayed cross-model feedback.

Overall, this adaptive rule maintains the efficiency of greedy decoding while
adjusting the degree of commitment based on model confidence, striking a
practical balance between accuracy, fluency, and computational cost.

\subsection{Diversity-Aware Ensemble Scaling}
\label{subsec:diversity-aware-ensemble-scaling}
Building on the adaptive word-length expansion strategy, \textsc{AdaFuse} explores the role of lexical diversity in word-level ensemble decoding.
To balance diversity and efficiency, we adopt an adaptive mechanism that expands the candidate space only when additional exploration is warranted, rather than enforcing uniform diversification.
Since beam search operates at the token level and prioritizes high-probability continuations, directly applying it at the word level often yields redundant candidates with limited effective diversity.
To address this mismatch, \textsc{AdaFuse} employs a two-stage word-level search that separates \emph{exploration} and \emph{exploitation}, enabling controlled diversity while preserving fluency.

\paragraph{Adaptive Trigger.}
Diversity is introduced only when the first-token distribution at the current decoding prefix $\mathbf{y}$ fails to satisfy the confidence criterion in Eq.~(\ref{eq:confidence-criterion}), indicating elevated uncertainty in initiating a new word.
In such cases, \textsc{AdaFuse} activates the two-stage diversity search to enumerate multiple plausible word-level continuations.
When the criterion is satisfied, decoding proceeds without diversification, ensuring that diversity is applied only when beneficial while maintaining computational efficiency.

\paragraph{Exploration.}
In the first stage, we select the top-$B$ \emph{distinct} initial tokens based on their conditional likelihood given the current decoding prefix $\mathbf{y}$.
We refer to $B$ as the \emph{branching factor}, which determines how many paths are explored per decoding step to encourage intra-round diversity:
\begin{equation}
\{z^{(1)}_1, \dots, z^{(B)}_1\} = \text{Top-}B \left( p(v \mid \mathbf{y}) \right),
\end{equation}

Each $z^{(b)}_1$ serves as a distinct lexical entry point for the generation of candidate words.

\paragraph{Exploitation.}
Each selected initial token $z^{(b)}_1$ is then greedily extended autoregressively to form a complete word:
\begin{equation}
\begin{aligned}
z^{(b)}_i
&= \arg\max_{v} \; p_{m_k}\!\left(
v \mid \mathbf{y}, z^{(b)}_1, \dots, z^{(b)}_{i-1}
\right), \\
&\qquad i = 2, 3, \dots
\end{aligned}
\end{equation}

Decoding continues until a whitespace character is produced, at which point we obtain the complete word sequence candidates:
$
\mathbf{w}^{(b)} = (z^{(b)}_1, \dots, z^{(b)}_L),
$
where $L$ denotes the position at which decoding yields a full word.
This two-stage procedure produces diverse yet fluent candidate words for subsequent ensemble scoring.

\subsection{Cross-Model Scoring and Candidate Integration}
\label{subsec:cross-model-scoring}
The final step in \textsc{AdaFuse} is to select the most suitable continuation from a pooled set of word-level span candidates constructed across models.
Within each decoding round, each model generates a span candidate by concatenating the words it commits in that round.
We denote the $b$-th span candidate as
$
\mathbf{s}^{(b)} = \big(\mathbf{w}^{(b)}_1, \dots, \mathbf{w}^{(b)}_{c}\big),
$
where $c \le M$ is the adaptive number of committed words in the round. When the confidence criterion remains satisfied throughout the round, the model produces a single span candidate.
Otherwise, diversity-aware ensemble scaling is triggered to enumerate $B$ alternative span candidates: 
$
\mathbf{s}^{(1)}, \dots, \mathbf{s}^{(B)}.
$

\textsc{AdaFuse} forms a candidate pool $S$ by taking the \emph{union} of $\mathbf{s}^{(b)}$ generated by all participating models.
To compare candidates fairly, we compute the normalized negative log-likelihood (NLL) of each word as its average token-level surprise. NLL reflects model confidence over each candidate word, enables fair comparison across variable-length segments, and supports stable aggregation across models.

For each model $m_k$, we compute the normalized negative log-likelihood (NLL) of a span candidate $\mathbf{s}^{(b)}$ as the average token-level surprise:
\begin{equation}
\text{NLL}_{m_k}(\mathbf{s}^{(b)})
= -\frac{1}{T_b} \sum_{t=1}^{T_b}
\log p_{m_k}\!\left(
z^{(b)}_t \mid \mathbf{y}, z^{(b)}_{<t}
\right),
\end{equation}

\noindent where $(z^{(b)}_1, \dots, z^{(b)}_{T_b})$ denotes the token sequence obtained by concatenating all tokens in the span $\mathbf{s}^{(b)}$, and $T_b$ is the total number of tokens in the span.

We then average these scores across the ensemble of $K$ models to obtain the fusion score:
\begin{equation}
F(\mathbf{s}^{(b)}) = \frac{1}{K} \sum_{k=1}^{K} \text{NLL}_{m_k}(\mathbf{s}^{(b)}),
\end{equation}

Finally, we select the best candidate by minimizing the fusion score:
\begin{equation}
\mathbf{s}^* = \arg\min_{\mathbf{s}^{(b)}} F(\mathbf{s}^{(b)}),
\end{equation}
and append it to the decoding prefix:
\begin{equation}
\mathbf{y} \leftarrow \mathbf{y} \Vert \mathbf{s}^*.
\end{equation}

This scoring mechanism enables \textsc{AdaFuse} to act as a dynamic feedback system, selecting candidates that are not only contextually appropriate for the current prefix but also jointly supported across diverse model predictions. The complete procedure is outlined in Algorithm~\ref{alg:AdaFuse}.

\begin{algorithm}[t]
\small
\caption{\textsc{AdaFuse} Decoding}
\label{alg:AdaFuse}
\begin{algorithmic}[1]
\Require Prompt $P$, models $\{m_k\}_{k=1}^K$, margin threshold $\tau_\Delta$, max words $M$, branching factor $B$
\State $\mathbf{y} \gets P$ \Comment{Committed decoding prefix}
\While{not Terminate} \Comment{until EOS or length limit}
  \State $S \gets \emptyset$ \Comment{Candidate span pool}
  \For{each model $m_k$}
    \State $\mathbf{s} \gets \emptyset$, $c \gets 0$
    \While{$c < M$}
      \State $\mathbf{w} \gets \textsc{GenWord}\!\left(m_k,\, \mathbf{y}\Vert\mathbf{s}\right)$\Comment{\S~\ref{sec:candidate_word_proposal}}
      \If{$\Delta_{m_k}(\mathbf{y} \Vert \mathbf{s}) \ge \tau_\Delta$} \Comment{\S~\ref{subsec:adaptive-word-commitment}}
        \State $\mathbf{s} \gets \mathbf{s} \Vert \mathbf{w}$,  $c \gets c + 1$

      \Else      \Comment{\S~\ref{subsec:diversity-aware-ensemble-scaling}}

        \State \mbox{$\{\mathbf{w}^{(b)}\}_{b=1}^{B} \gets
        \textsc{GenWord}\!\left(m_k,\, \mathbf{y}\Vert\mathbf{s},\, B\right)$}
        \State $\mathbf{s} \gets \mathbf{s} \Vert \{\mathbf{w}^{(b)}\}_{b=1}^{B}$

        \State \textbf{break}
      \EndIf
    \EndWhile
    \State $S \gets S \cup \{\mathbf{s}\}$
  \EndFor

  \For{each candidate $\mathbf{s} \in S$} \Comment{\S~\ref{subsec:cross-model-scoring}}
    \State $F(\mathbf{s}) \gets \frac{1}{K}\sum_{k=1}^{K}\text{NLL}_{m_k}(\mathbf{s})$
  \EndFor
  \State $\mathbf{s}^* \gets \arg\min_{\mathbf{s}\in S} F(\mathbf{s})$
  \State $\mathbf{y} \gets \mathbf{y} \Vert \mathbf{s}^*$
\EndWhile
\State \Return $\mathbf{y}$
\end{algorithmic}
\end{algorithm}

\section{Experiments}
In this section, we conduct a series of experiments to systematically evaluate the effectiveness, and design decisions of \textsc{AdaFuse}. Specifically, we aim to answer the following research questions:

\begin{itemize}[itemsep=-0.2em, topsep=-0.3em, leftmargin=*]
    \item \textbf{RQ1}: Does \textsc{AdaFuse} consistently enhance the performance of base models across a wide range of tasks?

    \item \textbf{RQ2}: How does the adaptive word-commitment mechanism in \textsc{AdaFuse} affect the effectiveness of word-level ensemble decoding? 

    \item \textbf{RQ3}: How does incorporating diversity-aware ensemble scaling in \textsc{AdaFuse} influence overall performance as the number of candidates increases?

    \item \textbf{RQ4}: How does \textsc{AdaFuse} balance efficiency and effectiveness, and what qualitative behaviors can be observed from example analyses?
\end{itemize}

\subsection{Setup}
\paragraph{Models.} We conduct our experiments using four recent and widely adopted open-source chat and instruction-tuned models:LLaMA-3.1-8B-Instruct \cite{grattafiori2024llama}, Mistral-7B-Instruct-v0.3 \cite{jiang2023mistral}, Qwen3-8B \cite{yang2025qwen3}, and InternLM3-8B-Instruct \cite{cai2024internlm2}. These models cover diverse, practical size regimes and represent the latest publicly released versions available during our experiments.

\begin{table*}[t]
  \centering
  \small
  \setlength{\tabcolsep}{6pt}

  \caption{
Evaluation results on six benchmarks for base models and LLM ensemble methods.
Baseline ensemble methods use a fixed LLaMA-3.1-8B-Instruct and
Mistral-7B-Instruct-v0.3 pair.
\textsc{AdaFuse} is evaluated with both a fixed base pair and an oracle top-2
base selection.
The overall best result is shown in \textbf{bold}, while the best result among
baseline ensemble methods is \underline{underlined}.
The improvement row reports the relative percentage gains of \textsc{AdaFuse} over the best-performing baseline method.
}
   \centering
   \hspace*{-0.3cm}
   \scalebox{0.93}{
  \begin{tabular}{lccccccc}
    \toprule
    \textbf{Model} & \textbf{NQ} & \textbf{SQuAD} & \textbf{TriviaQA} & \textbf{GSM8K} & \textbf{Flores En→De} & \textbf{Flores De→En} & \textbf{Avg} \\
    \midrule
    \addlinespace[0.5ex]
    \multicolumn{8}{c}{\textit{Base Models}} \\
    \addlinespace[0.1ex]
    \midrule
    \rowcolor{lightwine}
    Mistral-7B-Instruct-v0.3    & 32.05 & 83.49 & 77.27 & 61.71 & 29.99 & 40.64 & 54.19 \\
    \rowcolor{lightlavender}
    LLaMA-3.1-8B-Instruct       & 34.24 & 80.13 & 78.97 & 81.05 & 31.33 & 41.78 & 57.92 \\
    Qwen3-8B                    & 23.85 & 76.72 & 64.72 & 89.99 & 17.45 & 36.95 & 51.61 \\
    InternLM3-8B-Instruct       & 27.15 & 81.77 & 65.38 & 76.72 & 28.17 & 37.37 & 52.76 \\
    \addlinespace[1ex]
    \midrule
    \addlinespace[1ex]
    \multicolumn{8}{c}{\textit{LLM Ensemble Methods}} \\
    \addlinespace[0.1ex]
    \midrule
    \textsc{LLM-Blender}                 & 36.56 & 82.13 & 75.58 & 62.55 & 34.45 & 41.56 & 54.97 \\
    \textsc{DeepEn}                      & 37.42 & 71.51 & 73.85 & \underline{67.63} & \underline{37.56} & 42.33 & 55.05 \\
    \textsc{SweetSpan}                   & 37.59 & \underline{86.58} & 80.75 & 63.38 & 33.74 & \underline{42.85} & \underline{59.16} \\
    UniTE                                & \underline{38.95} & 82.17 & \underline{81.38} & 64.59 & 34.98 & 41.67 & 57.29 \\

    \addlinespace[1ex]
    \midrule
    \addlinespace[1ex]
    \multicolumn{8}{c}{\textit{Our Ensemble Method}} \\
    \addlinespace[0.1ex]
    \midrule
    \textbf{\textsc{AdaFuse} (Top-2 Base)}
    & \textbf{42.85} & \textbf{90.38} & \textbf{82.17} & \textbf{90.25} & \textbf{39.83} & \textbf{45.25} & \textbf{65.12} \\

    \textbf{\textsc{AdaFuse} (Fixed Base)}
    & \textbf{42.85} & 90.15 & \textbf{82.17} & 79.15 & \textbf{39.83} & \textbf{45.25} & 63.23 \\
    & (+10.01\%) & (+4.12\%) & (+0.97\%) & (+17.03\%) & (+6.04\%) & (+5.60\%) & (+6.88\%) \\
    \addlinespace[0.5ex]
    \bottomrule
  \end{tabular}}
  \label{tab:AdaFuse_results}
\end{table*}

\paragraph{Baselines.} We compare our method against four representative ensembling baselines. 1) \textbf{\textsc{LLM-Blender}} \cite{jiang2023llm} uses a reward model ( $\mathtt{PairRanker}$) and a fusion model ($\mathtt{GenFuser}$) to rerank and merge responses. 2) \textbf{\textsc{DeepEn}}\cite{huang2024ensemble} projects candidate outputs into a shared latent space using vocabulary alignment and relative representation modeling. 3)  \textbf{\textsc{SweetSpan}} \cite{xu2024hit} is a span-level ensemble method that selects from model-generated span candidates via perplexity-based scoring. 4)  \textbf{\textsc{UniTE}} \cite{yao2024determine} applies a top-\textit{k} union filtering strategy for robust vocabulary alignment and token selection.

\paragraph{Benchmarks.} We evaluate six benchmarks across three task categories. 
1) \textbf{Knowledge‑ intensive QA:} $\mathtt{NaturalQuestions}$ (5‑shot) \cite{kwiatkowski2019natural}, a large‑scale open‑domain QA dataset built from real Google search queries; $\mathtt{SQuAD}$ (5‑shot) \cite{rajpurkar2016squad}, reading‑comprehension questions over Wikipedia passages; and $\mathtt{TriviaQA}$ (5‑shot) \cite{joshi2017triviaqa}, trivia questions paired with evidence documents harvested from the web. 
2) \textbf{Arithmetic Reasoning:} $\mathtt{GSM8K}$ (4‑shot with chain‑of‑thought prompting) \cite{cobbe2021training}, grade‑school math word problems requiring multi‑step reasoning. 3) \textbf{Machine Translation:} $\mathtt{FLores}$ \cite{goyal2022flores} English(En)→German(De) (0‑shot) and German(De)→English(En) (0‑shot), covering translation quality on both high‑ and low‑resource language pairs.

\paragraph{Experimental Setup.} For the main two-model ensembling setup, we choose Mistral-7B-Instruct-v0.3 and LLaMA-3.1-8B-Instruct as our fixed pair, given their strong, complementary performance profiles. We also include Qwen3-8B and InternLM3-8B-Instruct to test scalability and cross-architecture robustness. In our experiments, the confidence threshold is set to $\tau_\Delta = 0.7$ (see Appendix~\ref{sec:exp-setup-hyperparams} for hyperparameter analysis). Within each decoding round, we cap the maximum number of consecutively committed words at $M=3$, and bound the total context length by a task-dependent limit of up to 512 tokens. Performance is evaluated using standard metrics for each task: exact match accuracy on NQ, SQuAD, and TriviaQA, accuracy on GSM8K, and spBLEU for machine translation. Unless otherwise specified, Section~\ref{subsec:diversity-aware-ensemble-scaling} is disabled in the main experiments to ensure a fair comparison under comparable computational budgets, and is examined separately in the ablation study. We additionally examine how \textsc{AdaFuse} scales with the number of base models by expanding the ensemble beyond the primary two-model setup; detailed results are reported in Appendix~\ref{sec:ensemble_size}.

\subsection{Main Results (RQ1)}

As shown in \autoref{tab:AdaFuse_results}, \textsc{AdaFuse} achieves an average score of 63.23 across six benchmarks, outperforming the strongest prior ensemble method (\textsc{SweetSpan}, 59.16) by 4.07 points (6.88\% relative improvement).

\textsc{AdaFuse} delivers particularly strong gains on knowledge-intensive QA tasks. Compared to the single model LLaMA-3.1-8B-Instruct, it improves performance by +8.61 pts on $\mathtt{NQ}$, +10.25 pts on $\mathtt{SQuAD}$, and +3.20 pts on $\mathtt{TriviaQA}$. It also consistently outperforms all competing ensemble methods on both translation benchmarks, demonstrating robust cross-task generalization.

On the arithmetic-reasoning benchmark $\mathtt{GSM8K}$, \textsc{AdaFuse} attains an accuracy of 79.15, slightly below LLaMA-3.1-8B-Instruct (81.05). We attribute this gap to the large performance disparity between the underlying base models. When ensembling the two strongest base models on $\mathtt{GSM8K}$, \textbf{Qwen3-8B} and \textbf{LLaMA-3.1-8B-Instruct}, \textsc{AdaFuse} achieves 90.25 accuracy, confirming that base-model compatibility plays a critical role in ensemble effectiveness. Notably, \textsc{AdaFuse} still surpasses all other baseline ensemble methods on $\mathtt{GSM8K}$ by at least +11.52 pts.

We further analyze performance by fusion granularity to contextualize these gains. 
Sample-level ensembling (\textsc{LLM-Blender}) provides only modest improvements (e.g., +2.32 pts on $\mathtt{NQ}$ and +3.12 pts on $\mathtt{Flores\ En\to De}$), indicating limited robustness when candidate quality is uniformly low. Span-level methods (\textsc{SweetSpan}) achieve strong gains on extractive QA (+6.45 pts on $\mathtt{SQuAD}$), but yield smaller improvements on open-domain QA and translation. Token-level approaches (\textsc{UniTE}) further improve open-domain QA (up to +4.71 pts on $\mathtt{NQ}$), yet underperform on translation and incur substantially higher computational overhead. In contrast, \textsc{AdaFuse} consistently delivers the largest gains across tasks (e.g., +8.61 pts on $\mathtt{NQ}$ and +8.50 pts on $\mathtt{Flores\ En\to De}$), striking a favorable balance between accuracy, robustness, and efficiency.

\subsection{Ablation Study (RQ2)}
\label{sec:ablation}

To assess the importance of adaptive word commitment in \textsc{AdaFuse}, we conduct an ablation study in which we replace the adaptive word commitment with fixed-length word commitment, while keeping all other components unchanged.

\begin{figure}[!t]
    \centering
    \includegraphics[width=0.95\linewidth]{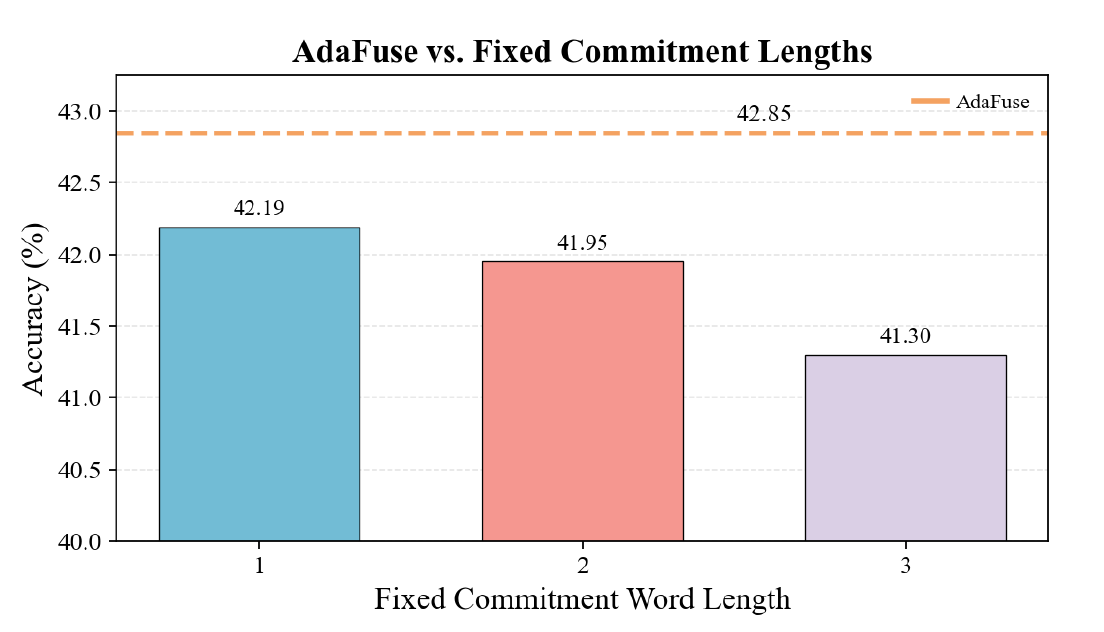}
    \caption{
        Comparison of Fixed-Length Decoding and \textsc{AdaFuse} on Natural Questions Dataset. 
    }
    \label{fig:AdaFuse_vs_Fixed}
\end{figure}

\begin{figure}[!t]
    \centering
    \includegraphics[width=1\columnwidth]{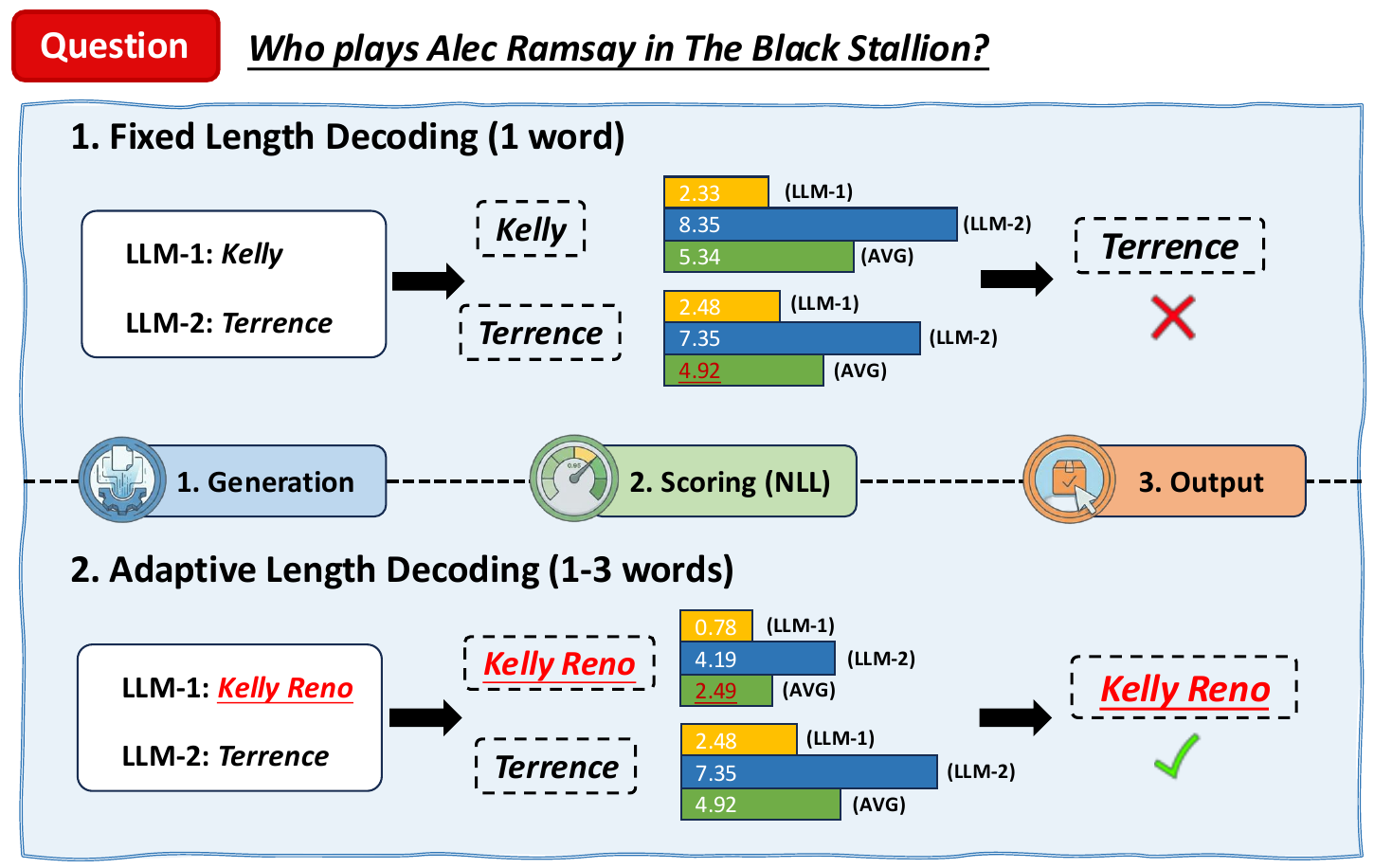}
    \caption{A Qualitative Example from Natural Questions: Fixed-Length Decoding vs. \textsc{AdaFuse}
}
    \label{fig:Example}
\end{figure}

\paragraph{Analysis of Adaptive Word Commitment.} We study the role of adaptive word commitment on the Natural Questions dataset using \textsc{AdaFuse} with the main fixed base-model pair. As shown in \autoref{fig:AdaFuse_vs_Fixed}, fixed-length decoding with word lengths of 1, 2, or 3 consistently underperforms adaptive \textsc{AdaFuse}. This is because fixed-length strategies tend to fragment semantically coherent units such as named entities and collocations, whereas adaptive commitment preserves complete semantic segments when confidence is high. Specifically, we select an example from the Natural Questions dataset and examine the same input question, as shown in \autoref{fig:Example}. With fixed-length decoding, the ensemble converges to an incorrect answer (“Terrence”). In contrast, adaptive word commitment allows one model to generate the correct multi-word entity (“Kelly Reno”) within a single decoding round, which is then selected by the subsequent cross-model scoring step.

\paragraph{Distribution of Words per Decoding Round.} We further analyze the distribution of the number of words generated per decoding round across different datasets. 
On Natural Questions, the proportions of rounds generating 1, 2, and 3 words are [80.29, 15.53, 4.18]\%, respectively; 
on SQuAD, they are [57.64, 21.27, 21.09]\%; 
and on De$\rightarrow$En translation, they shift to [34.02, 19.89, 46.09]\%. These results indicate that \textsc{AdaFuse} adapts the length of word-level commitment to the characteristics of different tasks, rather than relying on a fixed commitment length. 

\subsection{Scaling Analysis (RQ3)}
\begin{figure}[t]
    \centering
    \hspace*{-0.3cm}
    \includegraphics[width=1.1\linewidth]{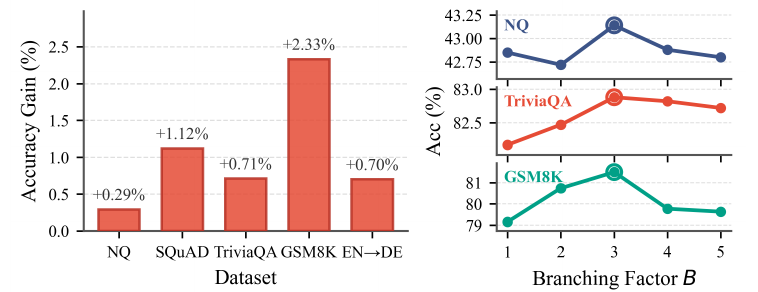}
    \caption{Diversity-Aware Ensemble Scaling Analysis
    }
    \label{fig:Diversity}
\end{figure}

We further examine the impact of introducing diversity-aware ensemble scaling into \textsc{AdaFuse}. Using two main models, we evaluate how adaptive branching influences performance across multiple datasets.
As shown in \autoref{fig:Diversity} (right), increasing the branching factor B from 1 to 5 on NQ, TriviaQA, and GSM8K leads to a generally upward trend in accuracy, suggesting that moderate branching helps the decoder explore more informative candidate paths. We then fix B=3 and compare \textsc{AdaFuse} with and without diversity-aware ensemble scaling across datasets. As illustrated in \autoref{fig:Diversity} (left), enabling diversity-aware ensemble scaling consistently improves performance on five benchmarks. These results indicate that, despite additional computational cost, \textsc{AdaFuse} provides a reliable accuracy gain by enhancing diversity during decoding.

In  Appendix~\ref{sec:appendix_beam}, we compare diversity-aware ensemble scaling with a beam-search-based scaling variant that ignores diversity in Figure~\ref{fig:Beamsearch}. The latter leads to a clear performance drop, while our approach remains stable and effective as the number of candidates increases.

\subsection{Case Study (RQ4)}
\begin{figure}[t]
    \centering
    \includegraphics[width=\linewidth]{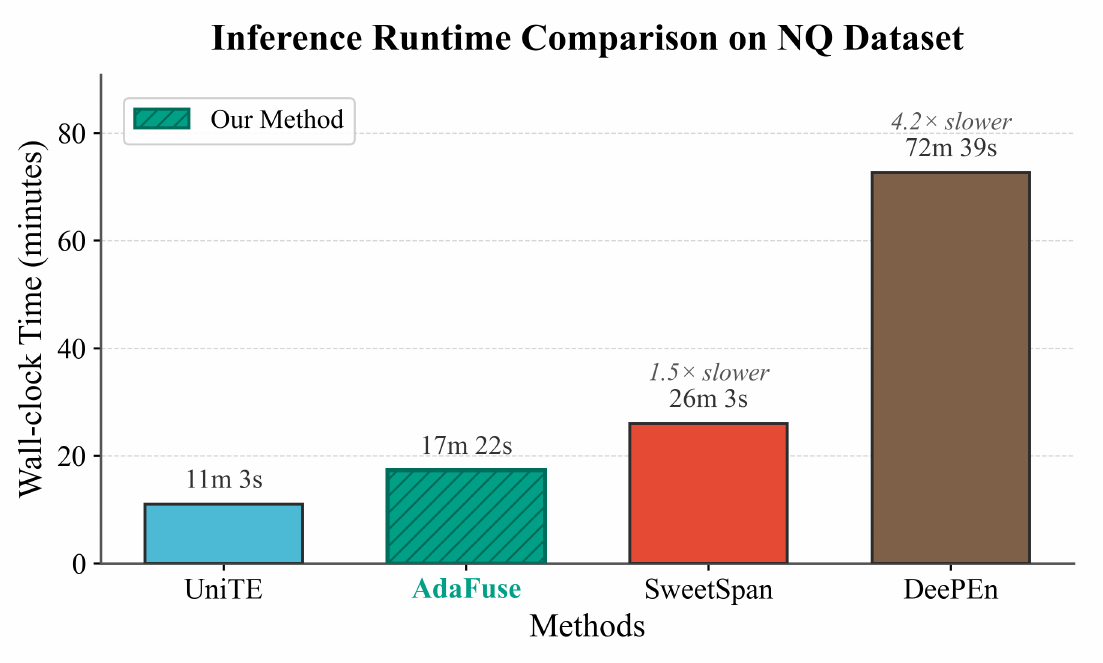}
    \caption{End-to-end runtime comparison of ensemble decoding methods on the Natural Questions (NQ) dataset under a unified multi-GPU inference setting.}
    \label{fig:runtime}
\end{figure}

\paragraph{Runtime Analysis.} Figure~\ref{fig:runtime} reports wall-clock inference time for different ensemble decoding strategies under a standardized setting (batch size 1, up to 10 new tokens per decoding call) on $4 \times$ NVIDIA A100 80GB GPUs using the main fixed base-model pair. \textsc{AdaFuse} achieves runtime performance comparable to UniTE, which relies primarily on lightweight tokenizer-level transformations and introduces minimal per-step overhead. At the same time, \textsc{AdaFuse} is substantially faster than span-level methods such as SweetSpan, whose fixed-span generation and scoring incur higher latency, and also outperforms token-level approaches like DeepEn that suffer from dense vocabulary alignment and cross-space projection costs. Overall, \textsc{AdaFuse} strikes a favorable balance between efficiency and quality, delivering improved accuracy while maintaining practical inference efficiency.
\paragraph{Case Analysis.} To better characterize the behavior of \textsc{AdaFuse} in longer-form generation, we additionally present a concrete generation example from the De$\rightarrow$En dataset in Example \ref{ex:translation}. In this example, each parenthesized span corresponds to one decoding round. As shown, \textsc{AdaFuse} frequently commits semantically cohesive word groups in longer sequences, reflecting its tendency to adapt word-level commitment to meaningful lexical units.

\vspace{4pt}
\refstepcounter{example}
\begin{examplebox}{Example~\theexample: De$\rightarrow$En Translation Instance}
\label{ex:translation}
\textbf{Source (De):}
Singapur ist im Allgemeinen ein äußerst sicherer Ort, an dem man sich gut zurechtfindet und nach der Ankunft fast alles kaufen kann.

\smallskip
\textbf{Output (En):}
[Singapore] [is generally a] [very safe place]
[where] [you] [can] [easily get around]
[and buy] [almost everything] [after arrival].

\smallskip
\textbf{Note:} [ ] denotes a word span generated in each decoding round.
\end{examplebox}

\section{Conclusion}

We introduced \textsc{AdaFuse}, an adaptive ensemble decoding framework that enables confidence-guided fusion of multiple large language models at inference time. Experiments across open-domain question answering, arithmetic reasoning, and machine translation show that \textsc{AdaFuse} consistently outperforms strong ensemble baselines under comparable computational budgets. Further analyses demonstrate the effectiveness of adaptive word commitment and controlled diversity in improving robustness and scalability.

More broadly, \textsc{AdaFuse} highlights the value of adaptive control for coordinating heterogeneous models during generation, allowing ensemble behavior to respond flexibly to uncertainty across diverse tasks and evolving generation contexts.

\section*{Limitations}

Our approach assumes access to token-level likelihoods and tokenizer outputs for cross-model scoring, which may not be available in closed-source or black-box LLM APIs. As a result, the applicability of \textsc{AdaFuse} is currently limited to open or fully accessible models, and extending the framework to settings with restricted model interfaces remains an open direction for future work.

\bibliography{ref}
\clearpage

\appendix
\section{Potential Risks}
\label{sec:Runtime}
\textsc{AdaFuse} adopts an adaptive decoding design that may increase GPU energy consumption, primarily due to a higher number of forward passes during inference. Compared to standard decoding, the framework repeatedly performs forward evaluations to generate candidate word spans and to score them across models, which increases the cumulative computational workload on the GPU. Nonetheless, relative to \emph{fixed-length} word-level ensembling, \textsc{AdaFuse} can reduce the total number of decoding rounds by committing longer multi-word segments when confidence is high. As a result, the additional per-round computation is offset by fewer decoding rounds and synchronization steps, often leading to improved end-to-end throughput despite the increased number of forward passes.


\section{Use Or Create Scientific Artifacts}  
Our work builds directly on publicly available benchmarks and contributes novel code artifacts to the community. Specifically, we use standard QA and translation datasets (NaturalQuestions, SQuAD, TriviaQA, GSM8K, FLORES) for evaluation, without modifying their contents. In addition, we develop and release the \textsc{AdaFuse} codebase, providing in a well‐organized repository with clear module structure.

\subsection{Cite Creators Of Artifacts}
All externally sourced artifacts are cited to their original publications and repositories. The NaturalQuestions \cite{kwiatkowski2019natural}, SQuAD \cite{rajpurkar2016squad}, and TriviaQA \cite{joshi2017triviaqa} benchmarks are credited to their respective authors; the GSM8K dataset is attributed to \cite{cobbe2021training}; and the FLORES evaluation suite is acknowledged to its creators,\cite{goyal2022flores}. 

Each model family used (LLaMA‑3.1, Mistral‑7B, Qwen3‑8B, InternLM3‑8B) is referenced via its official technical report or HuggingFace model card, ensuring that credit is properly given to all upstream contributors.

\subsection{Discuss The License For Artifacts}
We adhere to the licenses of all artifacts we use or release. All question‑answering and translation benchmarks are distributed under Creative Commons or public‐domain terms as specified by their maintainers.

Model checkpoints carry their original open‑source licenses (for example, MIT for Qwen3 and CC‑BY‑NC for InternLM3), and our own code and generated data are released under the MIT license. License files are included in our repository and documented in the supplementary README.

\subsection{Artifact Use Consistent With Intended Use}
We confirm that our use of each dataset and pre‑trained model aligns with its intended scope. All benchmarks are used strictly for inference and evaluation, consistent with their terms of service, and no modified model weights are redistributed. Our released code operates in inference mode only and does not enable fine‐tuning or commercial redistribution of the original checkpoints.

\subsection{Data Contains Personally Identifying Info Or Offensive Content}
Though benchmarks include public text with personal names or colloquial language, we:
\begin{itemize}
  \item Restrict to publicly released benchmark passages only.
  \item Apply keyword filtering to remove or mask offensive content in our analysis summaries.
\end{itemize}

\subsection{Documentation Of Artifacts}
All released artifacts are accompanied by clear and sufficient documentation to support reproducibility and reuse. The codebase is organized into modular components, with each module containing inline comments that describe its functionality and key parameters.

\subsection{Use of AI Assistants}

AI assistants (e.g., ChatGPT) were used to support writing clarity, language refinement, and minor code-level assistance during the preparation of this manuscript. All technical contributions, including method design, experimental setup, implementation decisions, and result analysis, were conceived and conducted by the authors. The use of AI assistants did not influence the scientific claims or empirical conclusions reported in this work.

\section{Statistics For Data}
The following descriptive statistics summarize the datasets used in this study:
\subsection{Flores (English to German) \textnormal{(Machine Translation)}}
\begin{itemize}[noitemsep, topsep=0pt]
  \item Number of entries: 1012
  \item Average question length: 130.4 characters
  \item Average answer length: 152.0 characters
\end{itemize}

\subsection{Flores (German to English) \textnormal{(Machine Translation)}}
\begin{itemize}[noitemsep, topsep=0pt]
  \item Number of entries: 1012
  \item Average question length: 152.0 characters
  \item Average answer length: 130.4 characters
\end{itemize}

\subsection{GSM8K \textnormal{(Arithmetic Reasoning)}}
\begin{itemize}[noitemsep, topsep=0pt]
  \item Number of entries: 1319
  \item Average question length: 239.9 characters
  \item Average answer length: 272.3 characters
\end{itemize}

\subsection{NaturalQuestions \textnormal{(Open‐Domain Question Answering)}}
\begin{itemize}[noitemsep, topsep=0pt]
  \item Number of entries: 3610
  \item Average question length: 47.7 characters
  \item Average answer length: 24.7 characters
\end{itemize}

\subsection{SQuAD 2500 \textnormal{(Reading Comprehension QA)}}
\begin{itemize}[noitemsep, topsep=0pt]
  \item Number of entries: 2500
  \item Average question length: 877.5 characters
  \item Average answer length: 66.1 characters
\end{itemize}

\subsection{Wikipedia Test 6000 \textnormal{(Open‐Domain QA)}}
\begin{itemize}[noitemsep, topsep=0pt]
  \item Number of entries: 6000
  \item Average question length: 78.8 characters
  \item Average answer length: 267.9 characters
\end{itemize}

\section{Computational Experiments}
All computational experiments described in this work are fully reproducible and documented in both the main text and the supplementary materials.

\subsection{Model Size And Budget}
For each LLM used, we specify its total parameter count and approximate GPU memory footprint: LLaMA‑3.1‑8B, Qwen3‑8B, and InternLM3‑8B each contain roughly 8 billion parameters and require about 30GB of GPU memory in 16‑bit precision, while Mistral‑7B has 7 billion parameters. The total compute budget for all experiments is approximately 500 A100 GPU‑hours, as detailed in Section 4.

\subsection{Experimental Setup And Hyperparameters}
\label{sec:exp-setup-hyperparams}

We describe every experimental setting in Section~4 and Appendix~C. All experiments are conducted on NVIDIA A100 80 GB GPUs. Key hyperparameters include a confidence threshold $\tau_\Delta = 0.7$, a maximum of 3 consecutively committed words per decoding round, and a task-dependent context length limit of up to 512 tokens. 

In addition, we conducted a targeted sensitivity analysis on the confidence threshold $\tau_\Delta$, which governs the adaptive decision about word-commitment in \textsc{AdaFuse}.

\begin{figure}[ht]
    \centering
    \includegraphics[width=1.0\columnwidth]{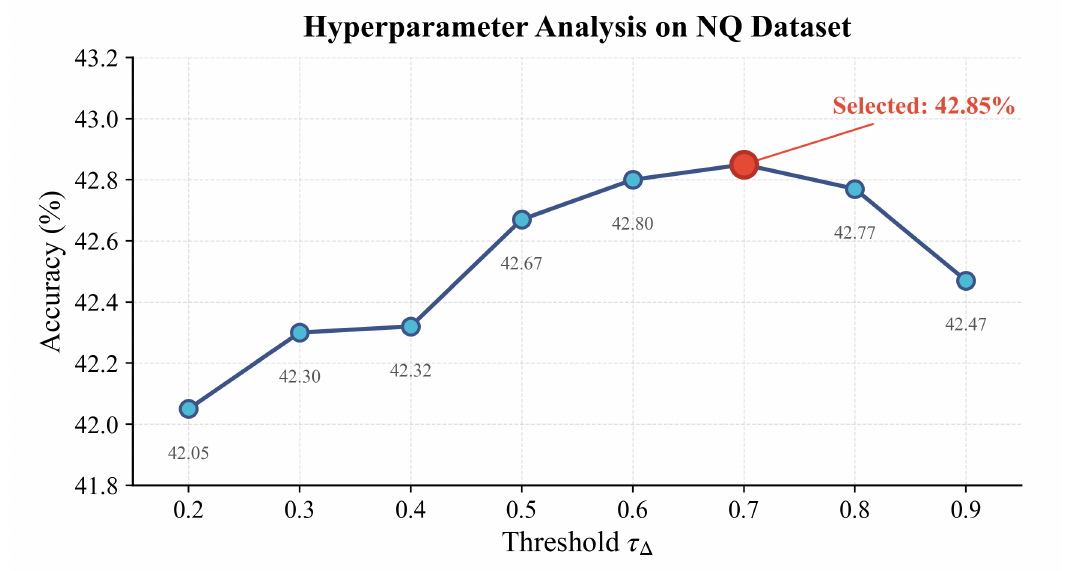}
    \caption{Effect of the confidence threshold $\tau_\Delta$ on \textsc{AdaFuse} performance on the Natural Questions dataset.}
    \label{fig:tau_analysis}
\end{figure}

As shown in Figure~\ref{fig:tau_analysis}, the performance of the model degrades noticeably when $\tau_\Delta$ is set to very small or very large values, indicating an increased sensitivity under extreme thresholds. In contrast, accuracy remains relatively stable in a broad region around $\tau_\Delta = 0.7$, with only minor fluctuations across nearby settings. Based on this observation, we adopt $\tau_\Delta = 0.7$ as a unified threshold for all datasets in our main experiments.

\subsection{Descriptive Statistics}
For each result table in the main text (Table~\ref{tab:AdaFuse_results}), we report results from a deterministic decoding setting.

\subsection{Parameters For Packages}
The software environment used in our experiments is fully specified in the supplementary materials. We use PyTorch (v2.4.1, CUDA 12.1) as the core deep-learning framework. Model loading, inference, and tokenization are implemented using HuggingFace Transformers (v4.51.3) and Tokenizers (v0.21.0).

For multi-GPU inference and distributed execution, we rely on Accelerate (v1.6.0). The data set handling and evaluation pipelines are implemented with the HuggingFace Datasets library (v3.0.2). 

Regarding evaluation, we report task-specific metrics consistent with prior work: exact match accuracy for question answering tasks (e.g., Natural Questions, SQuAD, TriviaQA), answer accuracy for GSM8K, and BLEU scores for machine translation benchmarks. In addition, we use the \texttt{BERTScore} package (v0.3.13) to compute BERTScore as a semantic similarity metric for generation quality assessment, where applicable.

\section{Effect of Ensemble Size}
\label{sec:ensemble_size}
We further examine how \textsc{AdaFuse} scales with the number of models in the ensemble on NQ dataset. 
Table~\ref{tab:ensemble_size} reports the Natural Questions accuracy obtained by incrementally adding models to the ensemble, starting from a two-model configuration and extending to three and four models.

The results show that \textsc{AdaFuse} achieves a clear improvement when moving from single-model baselines to a two-model ensemble. Adding a third model (Mistral-7B-Instruct-v0.3) further increases accuracy, indicating that incorporating a strong and complementary model can provide additional gains.  In contrast, extending the ensemble to four models by including Qwen3-8B leads to a slight performance drop, which can be attributed to the relatively weaker standalone performance of Qwen3-8B on NQ.

\begin{table}[t]
  \centering
  \scriptsize
  \setlength{\tabcolsep}{4pt}
  \caption{Impact of ensemble size on NQ accuracy.}
  \label{tab:ensemble_size}
  \resizebox{\columnwidth}{!}{%
  \begin{tabular}{l c}
    \toprule
    \textbf{Model \&\ Ensemble Setting} & \textbf{NQ Acc.} \\
    \midrule
    \rowcolor{lightwine}
    InternLM3-8B-Instruct              & 27.15 \\
    \rowcolor{lightlavender}
    LLaMA-3.1-8B-Instruct               & 34.24 \\
    Qwen3-8B                            & 23.85 \\
    Mistral-7B-Instruct-v0.3            & 32.05 \\
    \textbf{\textsc{AdaFuse}} (2 models, B=3) & 40.39\\ 
    +\,Mistral-7B-Instruct-v0.3 (3 models) & \textbf{42.63}  \\
    +\,Qwen3-8B (4 models)                 & 40.55 \\
    \bottomrule
  \end{tabular}}
\end{table}

\section{Diversity-Aware Scaling vs. Beam-Search Scaling}

\label{sec:appendix_beam}

\begin{figure}[!t]
    \centering
    \includegraphics[width=1.0\columnwidth]{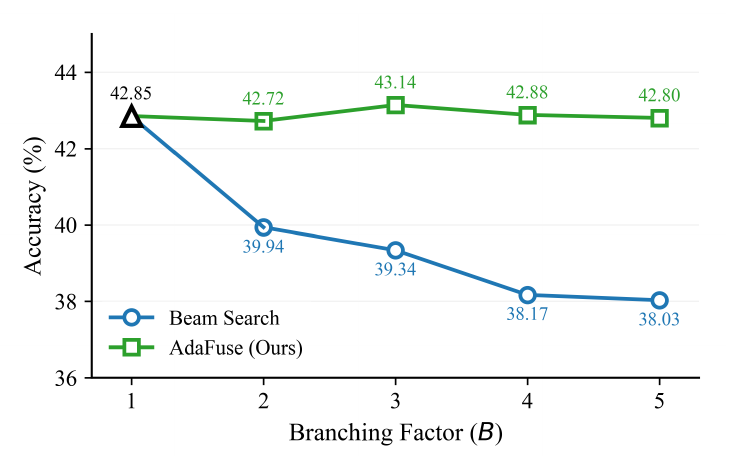}
    \caption{Comparison of Diversity-Aware Ensemble Scaling and Beam-Search-Based Scaling on the Natural Questions dataset.}

    \label{fig:Beamsearch}
\end{figure}

To further evaluate the effectiveness of diversity-aware ensemble scaling, we conduct an additional analysis on the Natural Questions (NQ) dataset using the same main fixed base-model pair as in the primary experiments, and compare \textsc{AdaFuse} with a beam-search-based scaling variant. Conventional beam search is formulated at the token level, while \textsc{AdaFuse} operates at the word level, where the number of subword tokens required to complete a word varies across lexical forms and tokenizers. This discrepancy prevents a direct alignment between beam-search steps and word-level decoding rounds. To enable a fair comparison, we implement a token-level beam search that generates a fixed number of new tokens per decoding round (set to 5 in our experiments), followed by decoding and truncation to the first complete word. As shown in \autoref{fig:Beamsearch}, this beam-search-based variant exhibits degraded scaling behavior as the branching factor increases. We attribute this outcome to the absence of adaptive word commitment and the lack of an explicit diversity-aware exploration--exploitation mechanism, which together result in highly redundant initial word candidates and limited effective diversity at the word level.

\end{document}